\documentclass{article}
\usepackage{spconf,amsmath,graphicx}
\usepackage{multirow, booktabs}
\usepackage{xcolor}
\usepackage{epstopdf}
\usepackage{subcaption}
\usepackage{graphicx}
\ninept

\title{Turn-taking and Backchannel Prediction with \\ Acoustic and Large Language Model Fusion}

\name{\parbox{\textwidth}{\centering
Jinhan Wang$^{1 \dagger}$ \qquad Long Chen$^{2 \dagger}$%
\thanks{$^{\dagger}$Equal contribution. This first author was an intern at Amazon.} \qquad Aparna Khare$^{2}$ \qquad Anirudh Raju$^{2}$ \qquad Pranav Dheram$^{2}$ \\
Di He$^{2}$ \qquad Minhua Wu$^{2}$ \qquad Andreas Stolcke$^{2}$ \qquad Venkatesh Ravichandran$^{2}$}}

\address{$^1$University of California, Los Angeles, USA \hspace{10mm}
      $^2$Amazon Alexa AI, USA}

\begin{document}
\ninept
\maketitle
\begin{abstract}
We propose an approach for continuous prediction of turn-taking and backchanneling locations in spoken dialogue by fusing a neural acoustic model with a large language model (LLM). Experiments on the Switchboard human-human conversation dataset demonstrate that our approach consistently outperforms the baseline models with single modality. We also develop a novel multi-task instruction fine-tuning strategy to further benefit from LLM-encoded knowledge for understanding the tasks and conversational contexts, leading to additional improvements. Our approach demonstrates the potential of combined LLMs and acoustic models for a more natural and conversational interaction between humans and speech-enabled AI agents.

\end{abstract}
\section{Introduction}
\label{sec:intro}

\label{sec:label}
%why not frame level
%1. self-defined heruistic variation
%2. frame-level has heavier computation burden on word-level

AI voice assistants are becoming increasingly multi-functional and important in people's daily lives \cite{hoy2018alexa,yaghoubzadeh2013virtual,rane2014study}. However, conventional voice assistant systems are mostly designed for query-based use cases. Towards the goal of more effective and effortless interaction between humans and AI, voice assistants that can solve tasks in a more natural manner and human-human-like conversational experience would be very desirable \cite{ward2015ten}. As one of its most basic capabilities, the system should be able to determine when to take turns naturally and with minimal latency in a dialogue with the user, and without the need for push-to-talk or wakewords. One common solution for turn-taking is to trigger the system's response after a period of silence based on a predefined threshold \cite{lala2018evaluation, chang2022turn, skantze2014turn, ward2010dialog}. However, this threshold-based method may result in a suboptimal user experience due to lack of naturalness \cite{aldeneh2018improving, ekstedt2022voice}. Another behavior that is important for managing human-human conversations that are a challenge for present-day conversational systems is backchanneling \cite{skantze2021turn,mueller2015using}. Backchannels are defined as short utterances expressing acknowledgment or reactions on the part of the listener, without signaling an intent to take a turn, such as ``\textit{uh-huh}", ``\textit{oh no}" and ``\textit{right}". They typically occur during the current speaker's turn and do not necessarily trigger turn-taking \cite{ortega2020oh, kawahara2016prediction, skantze2021turn}. 
% Thus it's of high ambiguity and challenges the model in decision making. 

Going back to conversation analysis in linguistic pragmatics \cite{sacks1978simplest}, there is a long history of descriptive and computational research trying to capture turn-taking and backchanneling cues in multiple modalities. In the acoustic domain, prosodic features such as duration, pitch, voice quality and intensity have been shown to have high correlation with turn-taking and backchannel locations \cite{skantze2017towards,skantze2021turn, liu2017turn, aldeneh2018improving}. In \cite{chang2022turn}, turn-taking prediction has been embedded as an auxiliary task in automatic speech recognition (ASR), based on acoustic encoder features. Aside from acoustic features, linguistic features have also been investigated. Given context or predicted transcription from an ASR system, word embeddings like Word2Vec \cite{lala2018evaluation} and encoded hidden states from transformer networks \cite{yang2022gated} or recurrent neural networks (RNNs) \cite{masumura2018neural, masumura2017online} have been used as linguistic representations for prediction. Furthermore, multi-modal fusion or joint modeling have been explored in earlier work, using RNN-based text and acoustic encoders \cite{roddy2018multimodal, liu2017turn}. 

However, in these earlier works that use linguistic modeling, features and representations are relatively simple and only approximate the full range of linguistic cues used by humans in daily conversation \cite{ekstedt2020turngpt}. Large language models (LLMs) promise to better capture the formal dependencies and meaning relations in language \cite{floridi2020gpt,jain2022jigsaw, kasneci2023chatgpt}.  Ekstedt et al.~\cite{ekstedt2020turngpt} proposed TurnGPT to leverage LLMs (in the form of GPT2 \cite{radford2019language}) for turn-taking prediction, showing superior performance compared to conventional modeling techniques. However, that work is still limited to turn-taking prediction and uses only lexical (text) information. 

In this work, we propose a novel approach for turn-taking and backchannel location prediction in spoken dialogue, with a fusion of LLM and acoustic models. We adopt two LLMs,  GPT2 \cite{radford2019language} and RedPajama \cite{together2023redpajama} for modeling linguistic cues, and we use HuBERT \cite{hsu2021hubert} for modeling acoustic cues, to leverage both representations and prior knowledge learned during pretraining. Two fusion methods are explored by manipulating the LLM branch to better understand the role of the different modalities in joint modeling. Furthermore, inspired by the success of instruction fine-tuning of LLMs for other tasks \cite{wei2021finetuned, min2021recent}, a novel multi-task instruction fine-tuning is proposed to further utilize the ability of LLMs  to understand task descriptions and dialogue history, and direct the joint model to focus on different tasks with task-specific submodules triggered by corresponding instructions. 
Our main contributions are thus (1) extending the turn-taking model to include backchanneling, (2) use of LLMs with acoustic fusion for these tasks, and (3) exploration of LLMs for instruction-tuning rather than simple token encoding and prediction.
%In contrast to previous works including TurnGPT, which either uses a single modality or simpler language models, our approach utilizes the fusion of LLM and acoustic model for turn-taking and backchannel prediction. Furthermore, instead of using the language model simply as a text encoder, we also explore the potential of LLMs to model context and learn from instruction-based multi-task fine-tuning. 

% Key points:

% Importance and challenge of conversational dialogue modeling task: more natural voice assistant system

% Acoustic and Linguistic approaches

% Advance in LLM systemd development

% Proposed methods

\section{Proposed Method}
\label{sec:format}
\subsection{Problem setup}
For a more natural human-agent interaction, the task of interest here is to predict the proper turn-related behavior with respect to the user's input to a voice assistant system during conversation. Three distinct behaviors are considered: 1) \textbf{Continuing Speech}: the currently active speaker is predicted to continue speaking (the other party keeps listening); 2) \textbf{Backchannel}: the listening party (system or user) should generate a brief utterance as a sign of acknowledgment, understanding or assessment without an intention to take the turn \cite{ortega2020oh}; 3) \textbf{Turn-taking}: the current speaker is predicted to be done talking and the nonspeaking party should take over the conversation and provide a response. More formally, given a (partial) utterance with acoustic features $X^A$ and text features $X^L$, the goal is to predict the class/behavior posteriors $P(Y|X^A,X^L)$, where $Y$ is from the label set consisting of ``Continuing Speech'', ``Backchannel'' and ``Turn-taking''. The framework is depicted in Figure~\ref{fig:diagram}.

\subsection{Acoustic and language modeling}
\label{ssec:acoustic_modeling}
The acoustic model is shown as the bottom left module in Figure~\ref{fig:diagram}. In this work, HuBERT \cite{hsu2021hubert} is used to encode speech signals of the (partial) utterances, as well as to serve as the base acoustic model (AM) for prediction with single modality. To manipulate the architecture for classification, the average-pooled 768-dimensional HuBERT embedding across all time steps emitted from the base model is fed into a projection layer to obtain a 256-dimensional vector. Then a linear classifier maps the projection to three classes. During model training, the acoustic base model is frozen.

\label{ssec:llm_fine_tune}
The linguistic modeling is done by LLM fine-tuning as shown at the bottom right in Figure~\ref{fig:diagram}. Here, either GPT2 or RedPajama are used to encode the text of the (partial) utterances, producing embeddings of 768 and 2560 dimensions, respectively. 
%Two different sizes of LLMs are investigated to demonstrate the scalability of the proposed method.
Unlike in acoustic modeling, LLM fine-tuning uses the embedding of the last token. Then, the embedding is fed into a linear layer of dimension 3 for classification. Depending on the base LLM being used, different fine-tuning strategies are applied, as discussed further in
Section~\ref{ssec:expts_details}.

\subsection{Fusion or joint training}
\label{ssec:fusion}
\begin{figure}[t]
  \centering
  \includegraphics[width=0.5\textwidth]{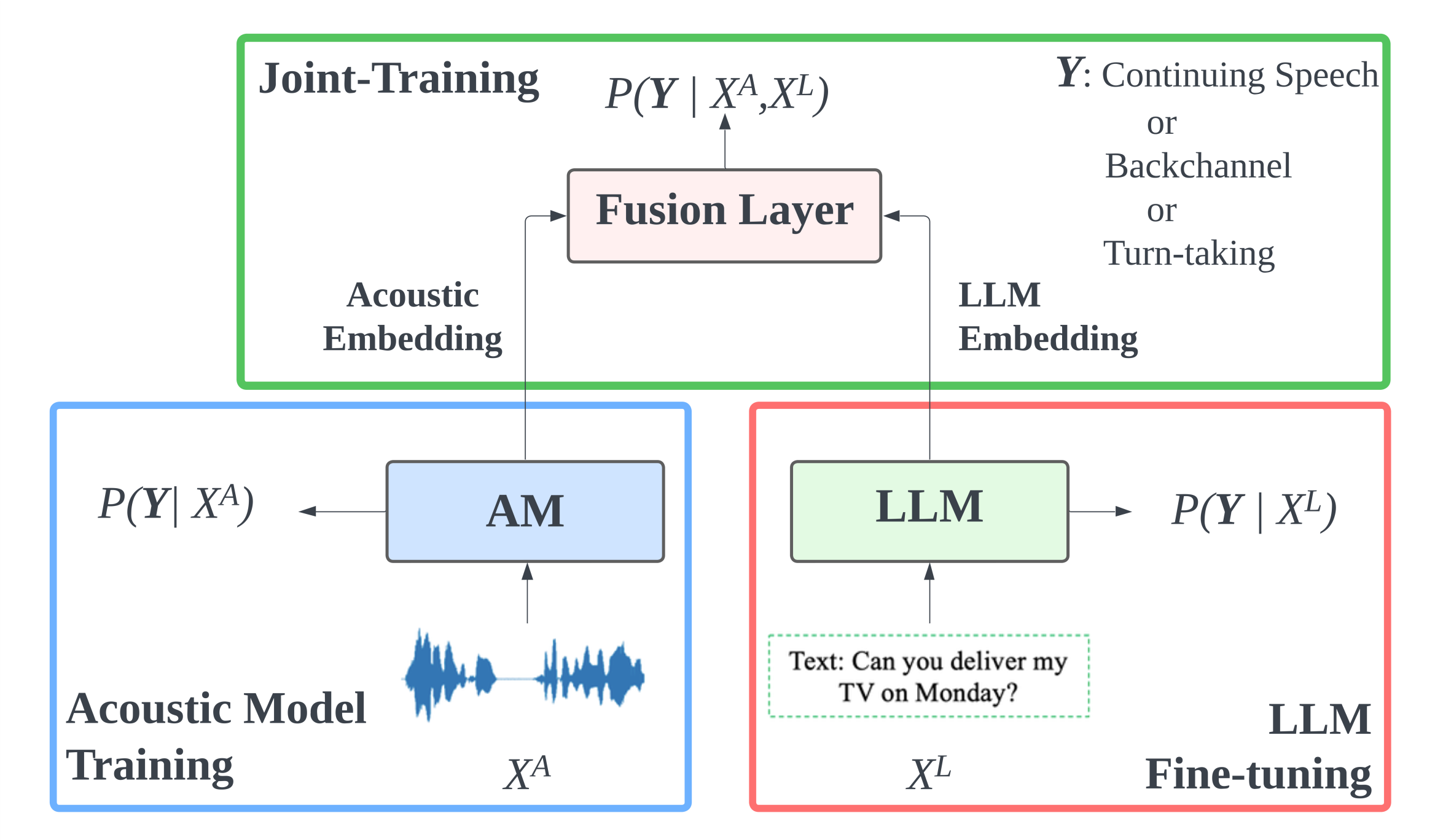}
    \vspace{-15pt}
  \caption{Schematic of combined acoustic and LLM modeling for turn-taking, backchannel and continuing speech prediction.}
  \label{fig:diagram}
\end{figure}

A late fusion mechanism is used where the final embeddings emitted from the AM and LLM are concatenated and fed into a single linear classification layer with dimension 3 to predict $P(Y|X^A,X^L)$, as shown in the top module of Figure~\ref{fig:diagram}. Two different fusion setups are investigated. In Option 1 (\textbf{Opt1}) both AM and LLM are loaded from the pretrained library \cite{wolf-etal-2020-transformers} without fine-tuning. Then, both the fusion layer and the LLM base model undergo domain adaptation and downstream task training. In Option 2 (\textbf{Opt2}), aside from loading the pretrained AM as in Opt1, the LLM is loaded after stand-alone fine-tuning as described in Section~\ref{ssec:llm_fine_tune}. Then the LLM branch is also frozen and only the fusion layer is trained. The key difference between \textbf{Opt1} and \textbf{Opt2} is whether LLM has been fine-tuned for the downstream task and frozen. Though more sophisticated architectures could be helpful, we will demonstrate the effectiveness of combining AM and LLM for turn-taking and backchannel prediction tasks even with the two simple fusion options considered here.

% It can be anticipated that a deeper and more intelligent post-processing layer module can help with the performance for both LLM, AM and fusion experiments. However, as a work of proof of concept on the effectiveness of joint modeling acoustic and linguistic information, further investigation on fusion module architecture is not conducted. 

\subsection{Multi-task instruction fine-tuning}
\label{ssec:inst_multi}
\begin{figure}[t]
  \centering
  \includegraphics[width=0.5\textwidth]{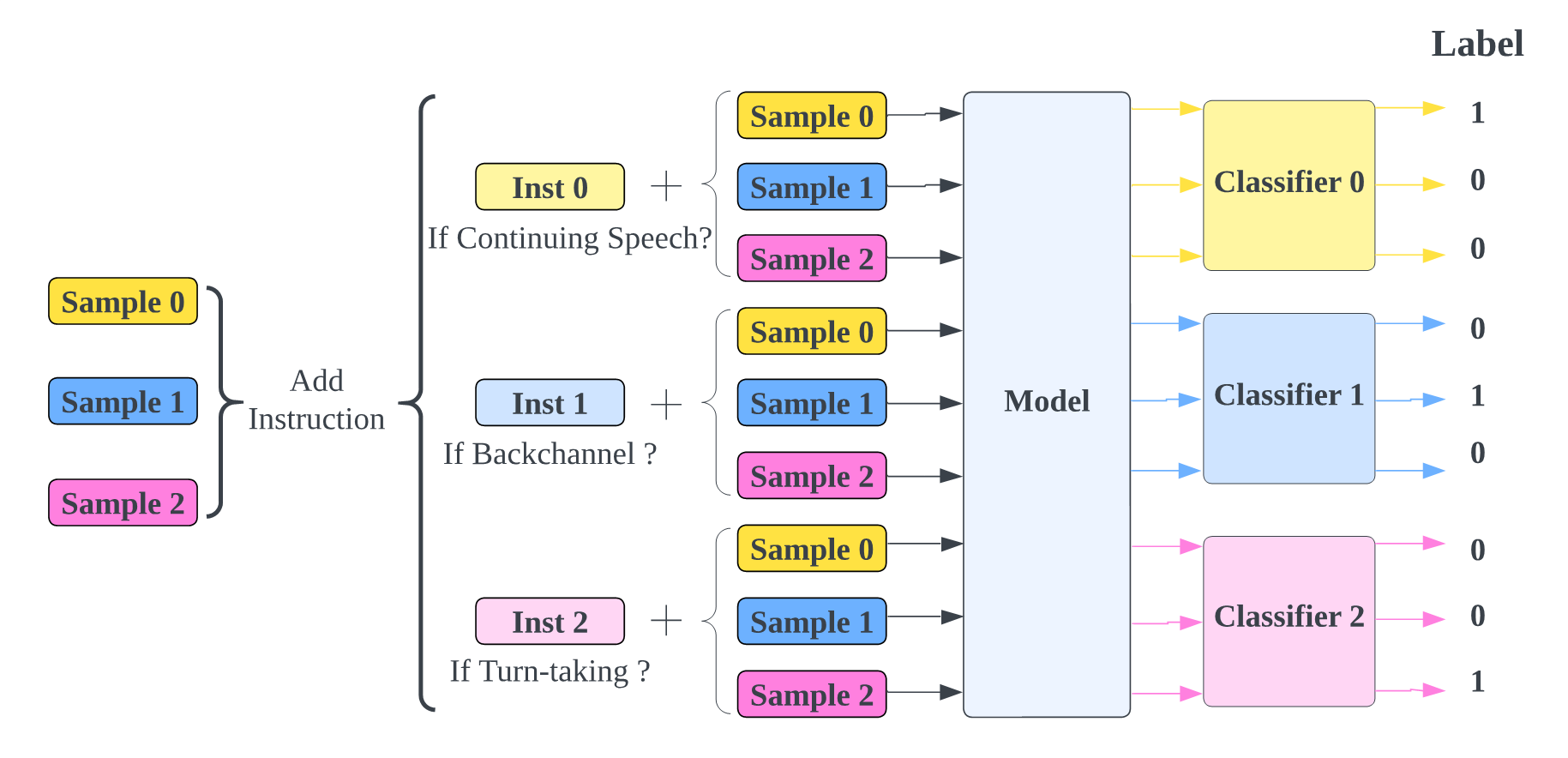}
    \vspace{-10pt}
  \caption{LLM-based multi-task instruction fine-tuning for turn-taking, backchannel, and continuing speech prediction.}
  \label{fig:inst}
\end{figure}

Besides serving as an advanced text encoder, LLMs have also demonstrated the ability to understand narrative instructions in natural language.  Instruction fine-tuning \cite{wei2021finetuned} has been used to teach LLMs this behavior. Thus, we reformulate our framework as a multi-task training scenario with instructions specific to our tasks. Rather than setting up a three-way classification, each class is handled as a separate binary classification task. This will later allows us to evaluate performance as three separate detection tasks. Figure~\ref{fig:inst} shows the diagram of this multi-task instruction fine-tuning process, where Sample 0, 1 and 2 are considered as the samples with corresponding ground-truth labels of ``Continuing Speech'', ``Backchannel'' and ``Turn-taking'', respectively. During training, each sample will be augmented three times, with the following respective instructions: 1) \textbf{Inst 0}: ``Identify if the current speaker will continue to speak at the end of the sentence."; 2) \textbf{Inst 1}: ``Identify if another speaker will backchannel at the end of the sentence."; 3) \textbf{Inst 2}: ``Identify if another speaker will take the turn at the end of the sentence."

% \begin{itemize}
%     \item \textbf{Inst 0}: "Identify if the current speaker will continue to speak at the end of the sentence."
%     \item \textbf{Inst 1}: "Identify if another speaker will backchannel at the end of the sentence."
%     \item \textbf{Inst 2}: "Identify if another speaker will take the turn at the end of the sentence."
% \end{itemize}

For each generated sample, if the prepended instruction corresponds to the ground-truth label, i.e. \textit{\{inst0, sample0\}}, \textit{\{inst1, sample1\}} and \textit{\{inst2, sample2\}}, then the corresponding binary label will be assigned as 1, otherwise 0. Each classifier is only in charge of one corresponding instruction and updates only its parameters, without being affected by samples augmented by the other two instructions. Let $X$ denote a batch of samples $X = [x_1, x_2, ..., x_n]$, with corresponding ground-truth labels $Y = [y_1, y_2, ..., y_n]$, and denote by $s$ the instruction index. The workflow can be written as follows:
\begin{equation}\label{eq:model}
\begin{aligned}
     {X_s} &  = \{inst_s, X\} \\
         &     = (inst_s, x_1), (inst_s, x_2)... (inst_s, x_n) \; s = 0, 1, 2 \\
\end{aligned}
\end{equation}
\begin{equation}\label{eq:label}
\begin{aligned}
    Y_{s} = [y_{s,1}, y_{s,2}, ..., y_{s,n}], \;\;\;
    y_{s,i} =  &
    \begin{cases}
    1& s = y_i \\
    0 & s \neq y_i
    \end{cases} \\
\end{aligned}
\end{equation}
\begin{equation}\label{eq:foward}
\begin{aligned}
    \hat{Y_{s}} = \mathit{Classifier_s}(\mathit{Model}(X_s))
\end{aligned}
\end{equation}
\begin{equation}\label{eq:loss}
\begin{aligned}
    L_{s} = \mathit{BCELoss}(\hat{Y_s}, Y_s) ,\;\;\;     L = \sum_{s=0}^{2}L_{s}
\end{aligned}
\end{equation}

%Compared to Section \ref{ssec:llm_fine_tune}, the LLM is not only used as text encoder, but also for task understanding. Additionally, in this independent classifier setup, because instructions are provided to be binary and task-oriented, the number of binary classifiers aligns with number of disjoint labels of the task. This configuration enables an easy scalability of the method towards additional speaker activity by adding binary classifier without altering the backbone model.

Compared to Section \ref{ssec:llm_fine_tune}, the LLM is used not only as a text encoder, but also for instruction understanding, leveraging pretrained knowledge about the tasks. Furthermore, having independent task-specific binary classifiers enables scaling to additional speaker activity classes or multi-label tasks. The training setup fully utilizes all original samples for each task to update the corresponding classifier by prepending the appropriate instruction.

We also explore a variant of instruction fine-tuning with added dialogue history to contextualize the model's interpretation. Here, two sentences preceding the target partial utterance, with speaker changes marked, are appended to the task-specific instruction, using the following format: ``Identify $<$instruction text$>$: $<$history with speaker token$>$. $<$target sample with speaker token$>$.'' 

% On one hand, each classifier is performing a binary classification with respect to the corresponding label as positive and the rest two are aggregated as negative ones, i.e. given a sample prepended by \textit{Inst 0}, \textit{classifier 0} will predict the score of whether the sentence ends with a continuous speech. On the other hand, each classifier can be guaranteed to be updated by the whole original dataset prepended by the corresponding instruction. 

\section{Experiments}
\label{sec:exp}
\subsection{Dataset}
\label{ssec:dataset}
We use the Switchboard corpus \cite{godfrey1992switchboard} (\textit{Switchboard-1 Telephone Speech Corpus: Release 2}). It is comprised of 2438 dyadic conversational dialogues involving 543 male and female speakers, who were connected by phone to converse  about one of around 70 topics. The dataset consists of around 260 hours of audio with ground-truth transcripts comprising around 3 million words; word-level time alignments are also available. We use these symmetrical human-human dialogues to model appropriate system behavior for a conversational voice assistant when receiving user input. Though all data consists of human-human dialogue, we treat the currently active user's speech as if input to a dialogue system, and the other (listening) speaker's behavior as a model for how the system should behave. As discussed, the possible behaviors are ``let current speaker continue", ``produce backchannel",  or ``take a turn". To utilize the data fully, user and system identities are swapped at each speaker change, i.e., when speaker A is active, A is the user and the system will try to behave as speaker B, and then vice versa for the next turn.

Given ground-truth speaker-wise dialog transcripts and word alignments, we prepare the data in each session as follows: 1) Extract dialog sentences from each speaker, while simultaneously normalizing special annotations, including [silence]/[noise] removal, partial word completion, and mispronunciation correction, as in \cite{ekstedt2020turngpt}. 2) Mark isolated one-word or two-word phrases as backchannel candidates. Backchannels are considered to be the 20 most frequent one and two-word phrases, such as ``yeah", ``mmhmm" and ``oh okay", as summarized in \cite{ekstedt2020turngpt}. 3) Combine the two speakers' dialog sentences in start-time ascending order and break sentences into words, for the purpose of word-level labeling in the following steps. 4) Remove words marked as backchannels and save them in a candidate list along with their speaker, start-time and end-time attributes. 5) Mark all speaker changes as ``Turn-taking" at last word spoken by a speaker. 6) Insert backchannel candidates back into the original dialogue according to their start-times and mark the word spoken by the other speaker where backchanneling occurs as ``Backchannel". %Then tokens which are not marked as ``backchannel`` or ``turn`` are labeled as ``continuing speech``. 
If a word is marked by none of these two labels, the default label of ``Continuing Speech`` is assigned.
%Ideally the backchannel labels should be annotated by human annotator or teach model. However, without such resources, the way we prepare the backchannel labels is the most practical way and same as the TurnGPT work.

Note that overlapping speech is only recorded for backchannel utterances, but not for regular turns, which are serialized, in a way compatible with TurnGPT processing \cite{ekstedt2020turngpt}.  We leave the prediction of turn-taking with overlap \cite{shriberg2001observations} for future work. (While overlapping turns are not uncommon for human-human dialog, a polite AI agent might refrain from producing them.)
After this preparation, each sample is a (partial) utterance (audio, text, or both) spoken by a single speaker, with the class label given by the last word's label within the utterance. The data is split by session with train:validation:test ratio of 2000:300:138 \cite{ekstedt2020turngpt}.

% The task is formulated as a sequence classification problem, where the model receives a (partial) utterance (audio, text or both), and predict the behavior of the other speaker at the end of the given sentence. Each sample is a (partial) sentence spoken by a single speaker with the corresponding label as the last word's label within the sentence. The dataset is split according to sessions: Train: 2000, Validation: 300, Test: 138. 

% \subsubsection{Extract Dialog}
% \subsubsection{Extract Backchannel Candidates}
% \subsubsection{Generate Turn Label}
% \subsubsection{Inject Backchannel}

% \subsection{Baseline}
% \label{ssec:baseline}

\subsection{Training and evaluation scenarios}
\label{ssec:traing_scenarios}

During training, since ``Continuing Speech" is by far the majority class, a downsampling procedure is applied to samples of that class, such that that the label frequency equals the average number of ``Backchannel" and ``Turn-taking" samples. The resulting subset has ``Continuing Speech", ``Backchannel", and ``Turn-Taking" occurring with counts 71k vs.\ 56k vs.\ 86k in training, and 6k vs.\ 5k vs.\ 7k for validation, respectively. During evaluation, all samples are used without downsampling, i.e., samples of each class in the test session will be decoded regardless of the class imbalance. The test set has samples with ``Continuing Speech", ``Backchannel", and ``Turn-taking" with counts 123k, 2.3k and 3.2k, respectively. 

In TurnGPT \cite{ekstedt2020turngpt}, the balanced accuracy (bAcc) over true and false turn-shifts is used as the evaluation metric. Here, since we have formulated three binary detection tasks, we prefer performance metrics that are independent of class priors and operating points (thresholds), namely, area-under-the-curve (AUC) and equal error rate (EER), evaluated for each class separately and in average.
The metrics are based on decision scores that are given by the logits for the targeted class, after softmax normalization.
%For 3-way classification, for a specific class, only the logit corresponding to that class after the softmax layer will be considered as a positive score. For multi-task instruction fine-tuning experiments, the single positive class prediction score from each binary classifier is used. 
%Regarding inference latency, though the task is conducted as a last-word sequence classification task in an offline manner, prediction delay can be taken as an upper-bound for token-level case in real-time where each token takes $\sim$7ms and $\sim$28ms for GPT2 and RedPajama, respectively.

\subsection{Experimental details}
\label{ssec:expts_details}
 All frameworks are implemented using the Huggingface Transformers Library \cite{wolf-etal-2020-transformers} on 8 NVIDIA V100 GPUs. To validate the generalization of the proposed methods, two pretrained LLMs of different sizes are investigated, namely GPT2 (124M parameters) \cite{radford2019language} and RedPajama (3B parameters) \cite{together2023redpajama}. For GPT2, the entire model is unfrozen for LLM fine-tuning and fusion Option 1. For RedPajama, a parameter-efficient fine-tuning approach, LoRA \cite{hu2021lora} with a rank of 32, is applied in LLM fine-tuning and fusion Option 1, resulting in around 0.4\% ($\approx$10M) trainable parameters. All models are trained with learning rate
 $5 \times 10^{-5}$, number of epochs 5, and batch size 4. All other hyperparameters are set to the default values provided by the Transformer library \cite{wolf-etal-2020-transformers}. 
% It is found that most models involved in this work are hyper-parameter insensitive, with minor performance variation through hyper-parameter grid-search. Thus, further investigation on optimal hyper-parameteres are omitted. 

\section{Results}
\label{sec:results}
\subsection{Single modality versus fusion}

Experimental results for single modalities and the two fusion approaches are reported in Table~\ref{tab:fusion}. First, with single modalities, language models yield much better performances than acoustic models. Second, by comparing the text-only models based on LLMs, it turns out that even though fine-tuning RedPajama results in significantly fewer trainable parameter than fully fine-tuning GPT2, RedPajama still achieves comparable performance, with an average AUC of 0.8351, as compared to 0.8292 for GPT2. This result indicates that RedPajama, as a larger LLM, has higher efficiency and greater potential for modeling conversational dialogue, and benefits more from approaches that exploit the model's language understanding capabilities, such as the proposed instruction fine-tuning. 

Under the same LLM setup, both fusion approaches achieve significant improvements over models with single modality, as shown in Table~\ref{tab:fusion}. For instance, the fusion model with RedPajama + HuBERT + Opt1 achieves the best performance for all three classes with an average AUC of 0.8657, leading to relative improvements of 22.6\% and 3.67\% over the best acoustic and text single modality models, respectively. Moreover, by comparing the three classes, predicting ``Continuing Speech" and ``Turn-taking" benefits most from the fusion, while ``Backchannel" only shows a small improvement. This result aligns with known properties of these turn-management events, where ``Turn-taking" and ``Continuing speech" are strongly cued by intonation and duration features, whereas ``Backchannel" is possibly more related to syntactic and semantic information. In addition, the difference between Opt1 and Opt2 does not share the same pattern for GPT2 and RedPajama. RedPajama works better with Opt1. We suspect that GPT2 has relatively limited modeling capacity. Unfreezing the larger RedPajama model will learn information that better complements the acoustic information.

We aimed to compare our model with TurnGPT \cite{ekstedt2020turngpt}, where bAcc of 0.789 and 0.823 were reported for the ``Spoken" dataset (of which Switchboard makes up the majority) with training on "Assistant" and "Full" datasets, respectively. We obtained a similar bAcc of 0.8002 for the GPT2-only model when focusing only on ``Turn-taking" and ``Continuing Speech" samples. Though this is not an apples-to-apples comparison, as the evaluation samples could be different, it shows that our LLM is comparable to TurnGPT when we leave out backchanneling.  In this focused two-class evaluation, we obtain an improved bAcc of 0.8578 for the RedPajama + HuBERT + Opt1 fusion model, confirming the benefit of complementing lexical with acoustic information.

\begin{table}[tb]
\centering
\caption{Results for single modality and fusion models.}
\vskip -1em
\resizebox{\linewidth}{!}{
\label{tab:fusion}
\begin{tabular}{cccccc}
\hline
 Method    &   {AUC(Cont)} & AUC(Back) & AUC(Turn)  & AUC(avg)       & EER(avg)      \\
 \hline
HuBERT & 0.7323 & 0.6455   & 0.7401  & 0.7060 & 34.87 \\
 \hline
GPT2   & 0.8510  & 0.7744  & 0.8623  & 0.8292  & 24.47 \\
+ HuBERT Opt1  & 0.8783  & 0.7798  & 0.884  & 0.8474 & 22.63 \\
+ HuBERT Opt2 & 0.8778  & \textbf{0.7862} & 0.8859 & 0.8500 & 22.77 \\
\hline
RedPajama & 0.8629  & 0.7739  & 0.8685 & 0.8351 &23.60 \\
+ HuBERT Opt1 & \textbf{0.8992} & \textbf{0.7862} & \textbf{0.9116} & \textbf{0.8657} & \textbf{20.33} \\
+ HuBERT Opt2 & 0.8982 & 0.7743 & 0.9006 & 0.8577 & 21.57 \\
\hline
\end{tabular}
}
\end{table}

% In GPT2 + Hubert experiments, fusion option1 gives 2.19\% relative improvement over GPT2 solely, and 2.51\% using fusion option2. In RedPajama + Hubert experiments, the performance is improved by 3.66\% and 2.71\% relatively using fusion option1 and option2 compared to using only RedPajama model. However, the difference between fusion option1 and option2 on two different LLM setups do not share the same pattern. It is suspected that GPT2 has relatively limited modeling capability in extracting linguistic information which can be compensatory to Hubert embedding in acoustic modeling. Thus, whether freezing a GPT2 model fine-tuned through in-domain data or unfreezing a raw pre-trained one do not differ much. As for RedPajama + Hubert fusion experiments, though improvements are all minor, all three classes AUC's are improved when applying option1 over option2. We believe it is because unfreezing the larger RedPajama model will let the model learn better compensatory information in collaboration with acoustic information provided by Hubert embedding.

\subsection{Multi-task instruction fine-tuning}

% \usepackage{multirow}
% \usepackage{multirow}

% \begin{table}[]
% \centering
% \caption{instruct}
% \resizebox{\linewidth}{!}{
% \label{tab:inst}
% \begin{tabular}{cccccc}
%  Exp    &   {AUC(Cont)} & AUC(Back) & AUC(Turn)  & AUC(avg)       & EER(avg)      \\\hline\hline
% GPT2                    &  0.8416   & 0.7863 &   0.8582  & 0.8287     & 24.13    \\\hline
% + Hubert Opt1      & 0.8726   &0.7901 &   0.8766  & 0.8464   & 22.5     \\\hline
% + Hubert Opt2      &  0.8806   &0.7838 &  0.8890   & 0.8511    & 22.23      \\\hline\hline
% RedPajama               & 0.8668 & 0.8097 &   0.8796     & 0.8520      & 21.8    \\\hline
% + Hubert Opt1 &  0.9  & 0.8229 &  0.9127  & 0.8785     & 19.5    \\\hline
% + Hubert Opt2 &  0.8980   &0.8182 &  0.9129  & 0.8764   & 19.6     \\\hline
% \end{tabular}
% }
% \end{table}

\begin{table}[t!]
\centering
\caption{Results with multi-task instruction fine-tuning.}
\vskip -1em
\resizebox{\linewidth}{!}{
\label{tab:inst}
\begin{tabular}{cccccc}
\hline
 Method    &   {AUC(Cont)} & AUC(Back) & AUC(Turn)  & AUC(avg)       & EER(avg)      \\
 \hline
GPT2                    &  0.8416   & 0.7863 &   0.8582  & 0.8287     & 24.13    \\
+ HuBERT Opt1      & 0.8726   &0.7901 &   0.8766  & 0.8464   & 22.50     \\
+ HuBERT Opt2      &  0.8806   &0.7838 &  0.8890   & 0.8511    & 22.23      \\
\hline
RedPajama               & 0.8668 & 0.8097 &   0.8796     & 0.8520      & 21.80    \\
+ HuBERT Opt1 &  0.9000  & \textbf{0.8229} &  0.9127  & 0.8785     &19.50    \\
+ HuBERT Opt2 &  0.8980   &0.8182 &  0.9129  & 0.8764   & 19.60     \\
\hline
RedPajama + History               & 0.8747 & 0.8074 &   0.8912     & 0.8578      & 21.63    \\
+ HuBERT Opt1 &  \textbf{0.9029}  & 0.8184 &  \textbf{0.9197}  & \textbf{0.8803}     &\textbf{19.30}    \\
\hline
\end{tabular}
}
\end{table}

\begin{figure}[t!]
\begin{subfigure}{0.49\linewidth}
  \centering
  % include first image
  \includegraphics[width=0.99\linewidth]{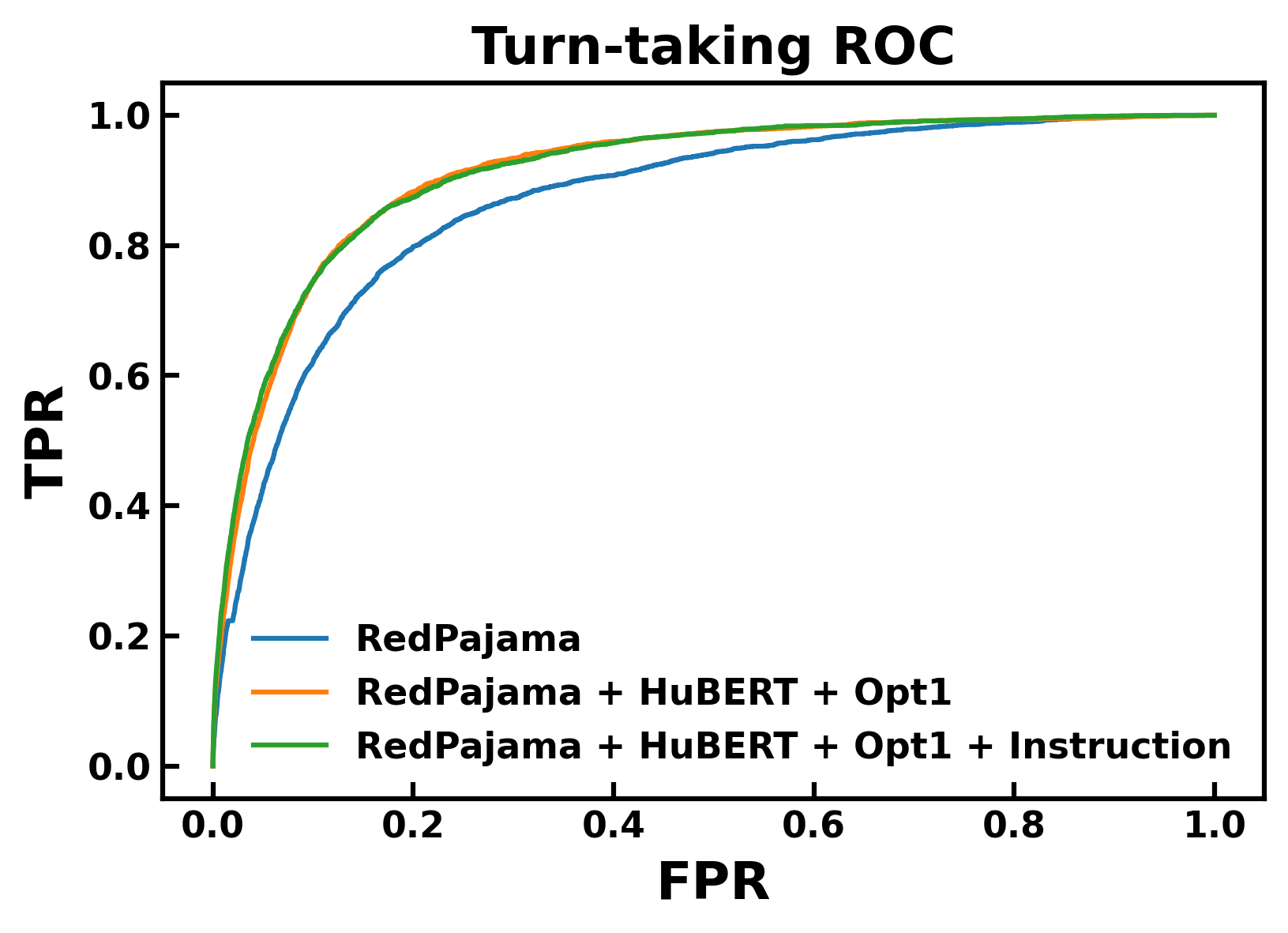}
  \vspace{-5pt}
  %\caption{ROC of turn-taking}
  \label{fig:sub-first}
\end{subfigure}
\begin{subfigure}{0.49\linewidth}
  \centering
  % include second image
  \includegraphics[width=0.99\linewidth]{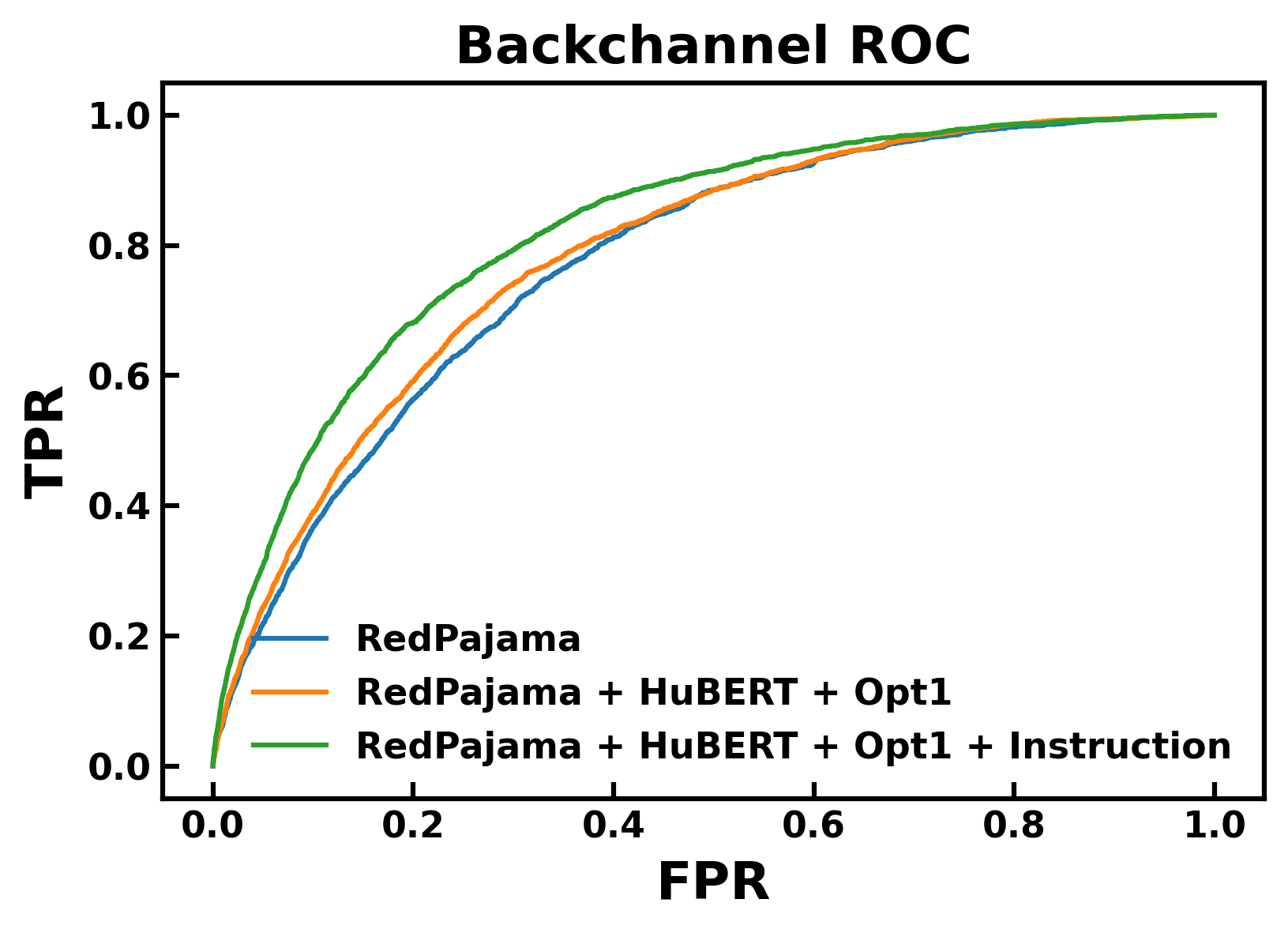} 
  \vspace{-5pt}
  %\caption{ROC of backchannel}
  \label{fig:sub-second}
\end{subfigure}
\vspace{-15pt}
\caption{ROC plots for turn-taking (left) and backchannel (right).}
\label{fig:roc}
\end{figure}

\begin{figure}[tb]
%\begin{subfigure}{0.5472\linewidth}
\begin{subfigure}[T]{0.49\linewidth}
  \centering
  % include first image
  \includegraphics[width=0.99\linewidth]{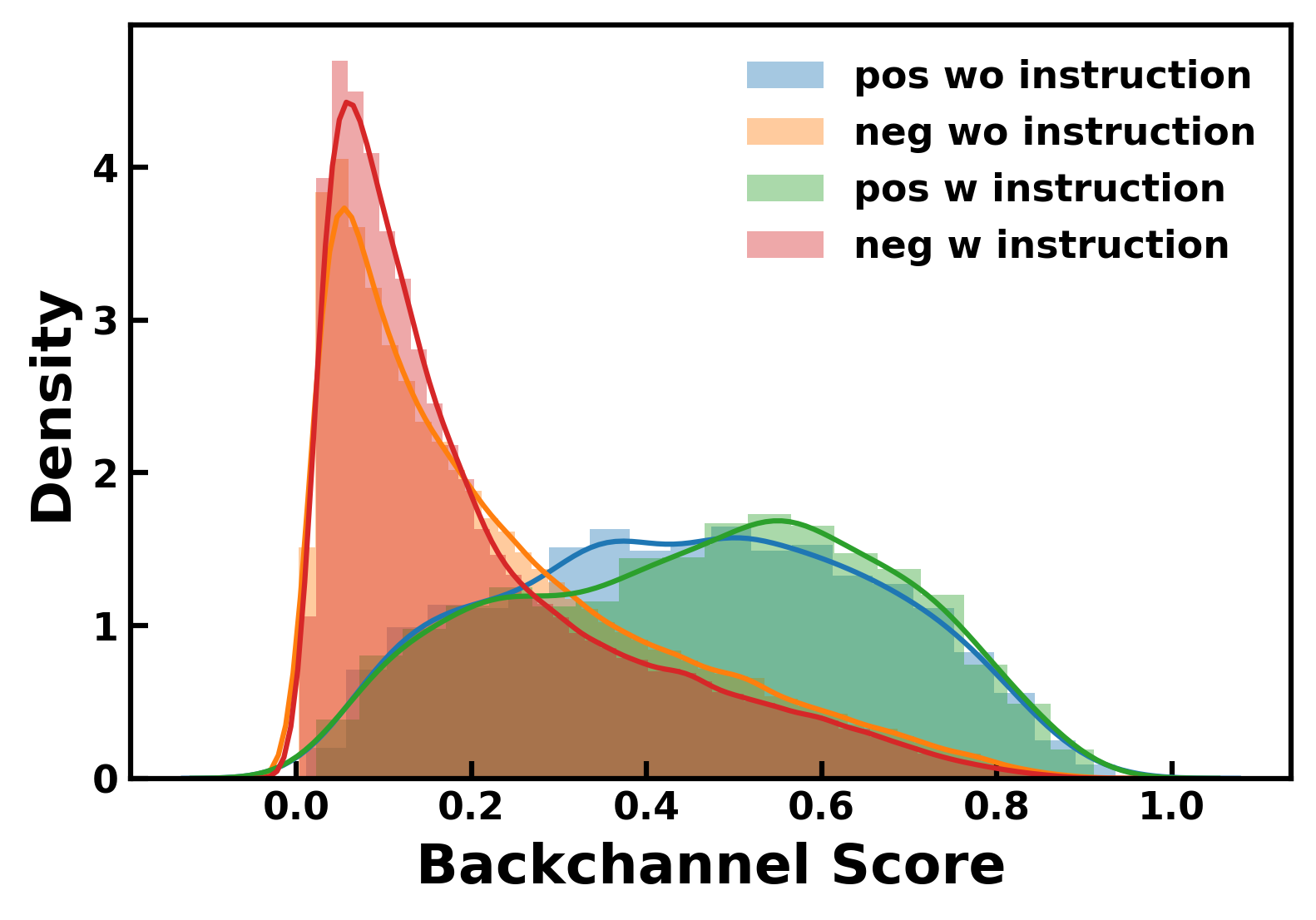}
  \vspace{-5pt}
  %\caption{Example 1}
  \label{fig:back_score_dist}
\end{subfigure}
%\begin{subfigure}{0.4328\linewidth}
\begin{subfigure}[T]{0.49\linewidth}
  \centering
  % include second image
  \includegraphics[width=0.99\linewidth]{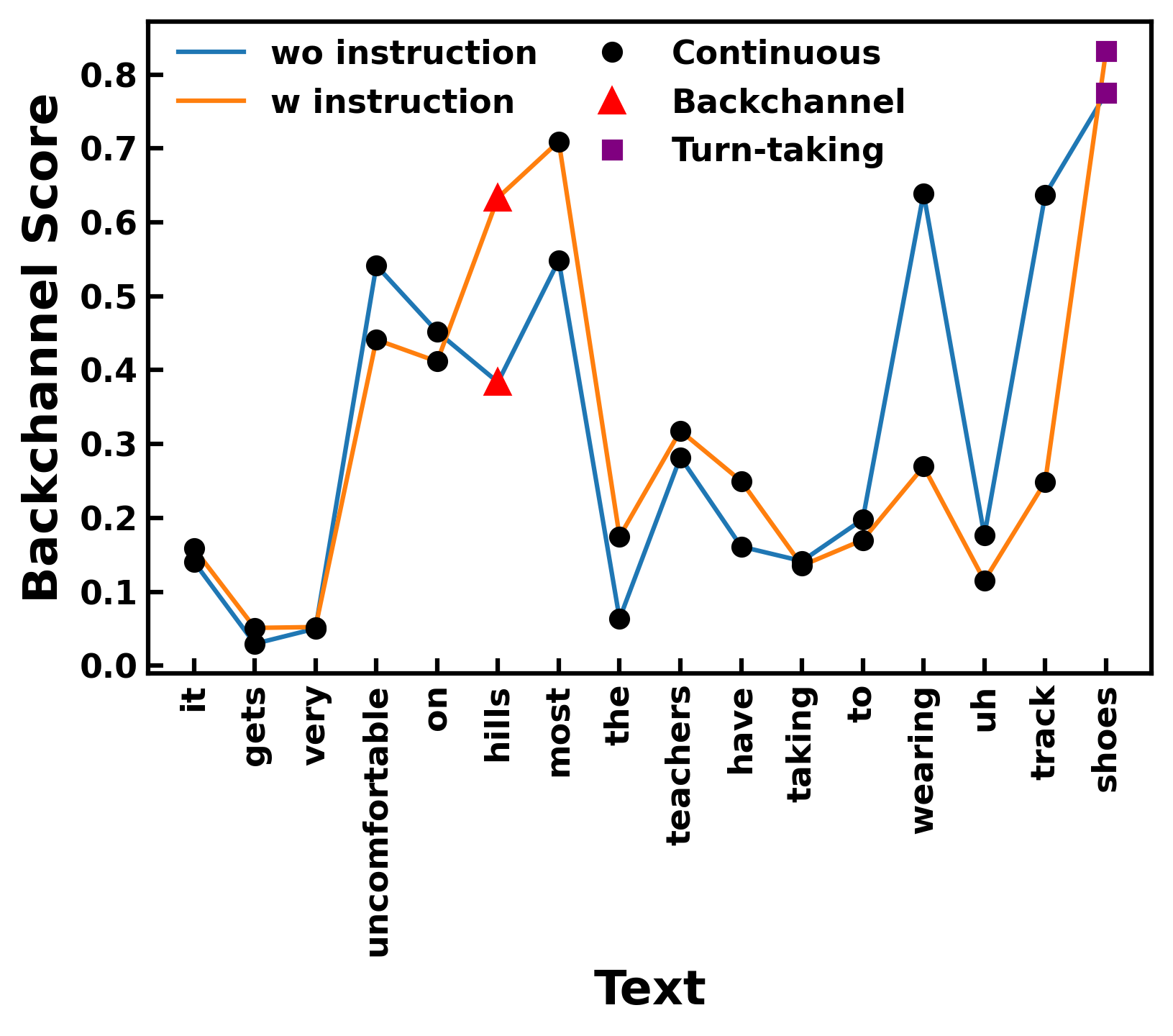} 
  \vspace{-5pt}
  %\caption{Example 2}
  \label{fig:back_examples}
\end{subfigure}
\vspace{-15pt}
\caption{Left: backchannel score distribution for the positive and negative samples. Right: a sentence example with token-level backchannel score. The markers represent the ground-truth token labels.}
\label{fig:back_analysis}
\end{figure}

Results for multi-task instruction fine-tuning are reported in Table~\ref{tab:inst}. For GPT2, it shows that applying instruction fine-tuning only results in a very minor differences to Table~\ref{tab:fusion}. However, when replacing GPT2 with RedPajama, significant improvements on average AUC and EER are observed for all three modeling approaches, with relative improvements of 2.02\%, 1.5\% and 2.16\% on average AUC for RedPajama, RedPajama + HuBERT Opt1 and Opt2, respectively. Moreover, RedPajama + HuBERT + Opt1 with multi-task instruction fine-tuning achieves the best performance for all the cases without the dialogue history, with an average AUC of 0.8785. 

More interestingly, comparing to the results  without the instruction fine-tuning in Table~\ref{tab:fusion}, ``Backchannel" prediction sees the highest AUC gain from applying the multi-task instruction fine-tuning, compared to other classes. Figure~\ref{fig:roc} shows the ROC curves for Turn-taking and Backchannel. It is clear here that Turn-taking benefits remarkably from the fusion, but benefits minimally from the instruction fine-tuning, while Backchannel shows the opposite trend. We conducted a further analysis by calculating the Backchannel score distribution of the samples. As shown in the left plot in Figure~\ref{fig:back_analysis}, applying instruction fine-tuning helps to push the score distribution of the backchannel (positive) samples and non-backchannel (negative) samples higher and lower, respectively. %We believe it is because the model trained with multi-task fine-tuning has a better sensitivity towards backchannel location. 
The right portion of Figure~\ref{fig:back_analysis} shows an example, with a transcript of ``It gets very uncomfortable on hills, most the teachers have taking%
\footnote{The provided corpus transcription. The actual word spoken is ``taken''.}
to wearing uh track shoes", the model with instruction fine-tuning correctly predicts the backchannel behavior after ``hills", while the model without predicts a backchannel after ``uncomfortable". This observation also supports our earlier conjecture that backchanneling is a speech activity best predicted by syntactic/semantic context. These results demonstrate that RedPajama benefits from multi-task instruction fine-tuning for better task understanding and more accurate backchannel prediction.

% These results clearly demonstrate that multi-task instruction fine-tuning can further leverage modern LLM for better task understanding and more accurate turn-taking prediction. 

\subsection{Instruction fine-tuning with dialogue history}

The last section of Table~\ref{tab:inst} shows the results for multi-task instruction fine-tuning with added dialogue history. As described earlier, a dialogue history of two sentences is included in the instruction, prepended to each sample utterance. Compared to the instruction fine-tuning results without history, average AUC and EER improve with history. However, when looking at each class individually, both ``Turn-taking'' and ``Continuing Speech'' classes are predicted better with history information, while the ``Backchannel'' class sees a slight degradation. This could be because backchanneling is largely a locally-cued behavior and affected little by long-term context.

% Key points to cover:

% Text-only: GPT2 vs RedPajama

% Acoustic Only: Hubert

% Fusion vs text vs acoustic

% Sequence vs Token-level training/evaluation

% Multi-task + instruction fine-tuning vs classification

% (optional) history/context

% (optional) frame-level

% (optional) text-to-text instruction fine-tuning

% To achieve the best rendering both in printed proceedings and electronic proceedings, we
% strongly encourage you to use Times-Roman font.  In addition, this will give
% the proceedings a more uniform look.  Use a font that is no smaller than nine
% point type throughout the paper, including figure captions.

% In nine point type font, capital letters are 2 mm high.  {\bf If you use the
% smallest point size, there should be no more than 3.2 lines/cm (8 lines/inch)
% vertically.}  This is a minimum spacing; 2.75 lines/cm (7 lines/inch) will make
% the paper much more readable.  Larger type sizes require correspondingly larger
% vertical spacing.  Please do not double-space your paper.  TrueType or
% Postscript Type 1 fonts are preferred.

% The first paragraph in each section should not be indented, but all the
% following paragraphs within the section should be indented as these paragraphs
% demonstrate.

\section{Conclusions}
\label{sec:majhead}

We have proposed a fusion model for turn-taking and backchannel prediction in spoken dialogue, combining both LLM and acoustic modeling. We experimented with LLMs of various sizes (GPT2 and RedPajama) and used HuBERT for modeling acoustic cues, to leverage both representations and prior knowledge learned from pretraining. Experiments demonstrate that our fusion approach consistently outperforms the baseline models with single modality, which indicates that joint modeling is effective at exploiting the complementarity of the modalities. Moreover, the proposed multi-task instruction fine-tuning strategy leverages LLMs for better task understanding and further improvements. Our approach provides a solution for more accurate causal turn-taking and backchannel prediction, ultimately enabling more natural and conversational human-agent interactions. In future work, it will be worth investigating how the use of automatic instead of ground-truth transcriptions would affect results, as required for inference in real-time applications.

\bibliographystyle{IEEEbib}
\footnotesize
\bibliography{strings,refs_new}

\begin{thebibliography}{10}

\bibitem{hoy2018alexa}
Matthew~B. Hoy,
\newblock ``Alexa, {Siri}, {Cortana}, and more: an introduction to voice
  assistants,''
\newblock {\em Medical reference services quarterly}, vol. 37, no. 1, pp.
  81--88, 2018.

\bibitem{yaghoubzadeh2013virtual}
Ramin Yaghoubzadeh, Marcel Kramer, Karola Pitsch, and Stefan Kopp,
\newblock ``Virtual agents as daily assistants for elderly or cognitively
  impaired people: Studies on acceptance and interaction feasibility,''
\newblock in {\em Intelligent Virtual Agents}, 2013, pp. 79--91.

\bibitem{rane2014study}
Pranav Rane, Varun Mhatre, and Lakshmi Kurup,
\newblock ``Study of a home robot: Jibo,''
\newblock {\em International journal of engineering research and technology},
  vol. 3, no. 10, pp. 490--493, 2014.

\bibitem{ward2015ten}
Nigel~G. Ward and David DeVault,
\newblock ``Ten challenges in highly-interactive dialog system.,''
\newblock in {\em AAAI Spring Symposium}, 2015.

\bibitem{lala2018evaluation}
Divesh Lala, Koji Inoue, and Tatsuya Kawahara,
\newblock ``Evaluation of real-time deep learning turn-taking models for
  multiple dialogue scenarios,''
\newblock in {\em Proc.\ ACM ICMI}, 2018, pp. 78--86.

\bibitem{chang2022turn}
Shuo-yiin Chang, Bo~Li, Tara~N Sainath, Chao Zhang, Trevor Strohman, Qiao
  Liang, and Yanzhang He,
\newblock ``Turn-taking prediction for natural conversational speech,''
\newblock in {\em Proc.\ Interspeech}, 2022, pp. 1821--1825.

\bibitem{skantze2014turn}
Gabriel Skantze, Anna Hjalmarsson, and Catharine Oertel,
\newblock ``Turn-taking, feedback and joint attention in situated human-robot
  interaction,''
\newblock {\em Speech Communication}, vol. 65, pp. 50--66, 2014.

\bibitem{ward2010dialog}
Nigel~G. Ward, Olac Fuentes, and Alejandro Vega,
\newblock ``Dialog prediction for a general model of turn-taking,''
\newblock in {\em Proc.\ Interspeech}, 2010, pp. 2662--2665.

\bibitem{aldeneh2018improving}
Zakaria Aldeneh, Dimitrios Dimitriadis, and Emily~Mower Provost,
\newblock ``Improving end-of-turn detection in spoken dialogues by detecting
  speaker intentions as a secondary task,''
\newblock in {\em Proc.\ ICASSP}, 2018, pp. 6159--6163.

\bibitem{ekstedt2022voice}
Erik Ekstedt and Gabriel Skantze,
\newblock ``Voice activity projection: Self-supervised learning of turn-taking
  events,''
\newblock in {\em Proc.\ Interspeech}, 2022, pp. 5190--5194.

\bibitem{skantze2021turn}
Gabriel Skantze,
\newblock ``Turn-taking in conversational systems and human-robot interaction:
  a review,''
\newblock {\em Computer Speech \& Language}, vol. 67, pp. 101178, 2021.

\bibitem{mueller2015using}
Markus Mueller, David Leuschner, Lars Briem, Maria Schmidt, Kevin Kilgour,
  Sebastian Stueker, and Alex Waibel,
\newblock ``Using neural networks for data-driven backchannel prediction: A
  survey on input features and training techniques,''
\newblock in {\em Human-Computer Interaction: Interaction Technologies}, 2015,
  pp. 329--340.

\bibitem{ortega2020oh}
Daniel Ortega, Chia-Yu Li, and Ngoc~Thang Vu,
\newblock ``Oh, jeez! or uh-huh? a listener-aware backchannel predictor on
  {ASR} transcriptions,''
\newblock in {\em Proc.\ IEEE ICASSP}, 2020, pp. 8064--8068.

\bibitem{kawahara2016prediction}
Tatsuya Kawahara, Takashi Yamaguchi, Koji Inoue, Katsuya Takanashi, and Nigel~G
  Ward,
\newblock ``Prediction and generation of backchannel form for attentive
  listening systems.,''
\newblock in {\em Interspeech}, 2016, pp. 2890--2894.

\bibitem{sacks1978simplest}
Harvey Sacks, Emanuel~A. Schegloff, and Gail Jefferson,
\newblock ``A simplest systematics for the organization of turn taking for
  conversation,''
\newblock in {\em Studies in the organization of conversational interaction},
  pp. 7--55. Elsevier, 1978.

\bibitem{skantze2017towards}
Gabriel Skantze,
\newblock ``Towards a general, continuous model of turn-taking in spoken
  dialogue using {LSTM} recurrent neural networks,''
\newblock in {\em Proceedings of the 18th Annual SIGdial Meeting on Discourse
  and Dialogue}, 2017, pp. 220--230.

\bibitem{liu2017turn}
Chaoran Liu, Carlos~Toshinori Ishi, and Hiroshi Ishiguro,
\newblock ``Turn-taking estimation model based on joint embedding of lexical
  and prosodic contents.,''
\newblock in {\em Proc.\ Interspeech}, 2017, pp. 1686--1690.

\bibitem{yang2022gated}
Jiudong Yang, Peiying Wang, Yi~Zhu, Mingchao Feng, Meng Chen, and Xiaodong He,
\newblock ``Gated multimodal fusion with contrastive learning for turn-taking
  prediction in human-robot dialogue,''
\newblock in {\em Proc.\ ICASSP}, 2022, pp. 7747--7751.

\bibitem{masumura2018neural}
Ryo Masumura, Tomohiro Tanaka, Atsushi Ando, Ryo Ishii, Ryuichiro Higashinaka,
  and Yushi Aono,
\newblock ``Neural dialogue context online end-of-turn detection,''
\newblock in {\em Proceedings of the 19th Annual SIGdial Meeting on Discourse
  and Dialogue}, 2018, pp. 224--228.

\bibitem{masumura2017online}
Ryo Masumura, Taichi Asami, Hirokazu Masataki, Ryo Ishii, and Ryuichiro
  Higashinaka,
\newblock ``Online end-of-turn detection from speech based on stacked
  time-asynchronous sequential networks.,''
\newblock in {\em Proc.\ Interspeech}, 2017, pp. 1661--1665.

\bibitem{roddy2018multimodal}
Matthew Roddy, Gabriel Skantze, and Naomi Harte,
\newblock ``Multimodal continuous turn-taking prediction using multiscale
  {RNNs},''
\newblock in {\em Proc.\ ACM ICMI}, 2018, pp. 186--190.

\bibitem{ekstedt2020turngpt}
Erik Ekstedt and Gabriel Skantze,
\newblock ``{T}urn{GPT}: a transformer-based language model for predicting
  turn-taking in spoken dialog,''
\newblock in {\em Proc. \ EMNLP}, 2020, pp. 2981--2990.

\bibitem{floridi2020gpt}
Luciano Floridi and Massimo Chiriatti,
\newblock ``{GPT-3}: Its nature, scope, limits, and consequences,''
\newblock {\em Minds and Machines}, vol. 30, pp. 681--694, 2020.

\bibitem{jain2022jigsaw}
Naman Jain, Skanda Vaidyanath, Arun Iyer, Nagarajan Natarajan, Suresh
  Parthasarathy, Sriram Rajamani, and Rahul Sharma,
\newblock ``Jigsaw: Large language models meet program synthesis,''
\newblock in {\em Proc.\ ACM ICSE}, 2022, pp. 1219--1231.

\bibitem{kasneci2023chatgpt}
Enkelejda Kasneci, Kathrin Se{\ss}ler, Stefan K{\"u}chemann, Maria Bannert,
  Daryna Dementieva, Frank Fischer, Urs Gasser, Georg Groh, Stephan
  G{\"u}nnemann, Eyke H{\"u}llermeier, et~al.,
\newblock ``{ChatGPT} for good? {On} opportunities and challenges of large
  language models for education,''
\newblock {\em Learning and individual differences}, vol. 103, pp. 102274,
  2023.

\bibitem{radford2019language}
Alec Radford, Jeff Wu, Rewon Child, David Luan, Dario Amodei, and Ilya
  Sutskever,
\newblock ``Language models are unsupervised multitask learners,''
\newblock {\em OpenAI Blog}, vol. 1, no. 8, 2019.

\bibitem{together2023redpajama}
Together Computer,
\newblock ``{RedPajama}: An open source recipe to reproduce {LLaMA} training
  dataset,'' https://github.com/togethercomputer/RedPajama-Data, Apr. 2023.

\bibitem{hsu2021hubert}
Wei-Ning Hsu, Benjamin Bolte, Yao-Hung~Hubert Tsai, Kushal Lakhotia, Ruslan
  Salakhutdinov, and Abdelrahman Mohamed,
\newblock ``{HuBERT}: Self-supervised speech representation learning by masked
  prediction of hidden units,''
\newblock {\em IEEE/ACM Transactions on Audio, Speech, and Language
  Processing}, vol. 29, pp. 3451--3460, 2021.

\bibitem{wei2021finetuned}
Jason Wei, Maarten Bosma, Vincent Zhao, Kelvin Guu, Adams~Wei Yu, Brian Lester,
  Nan Du, Andrew~M Dai, and Quoc~V Le,
\newblock ``Finetuned language models are zero-shot learners,''
\newblock in {\em Proc.\ ICLR}, 2022.

\bibitem{min2021recent}
Bonan Min, Hayley Ross, Elior Sulem, Amir Pouran~Ben Veyseh, Thien~Huu Nguyen,
  Oscar Sainz, Eneko Agirre, Ilana Heintz, and Dan Roth,
\newblock ``Recent advances in natural language processing via large
  pre-trained language models: A survey,''
\newblock {\em ACM Computing Surveys}, 2023.

\bibitem{wolf-etal-2020-transformers}
Thomas Wolf, Lysandre Debut, Victor Sanh, Julien Chaumond, Clement Delangue,
  Anthony Moi, Pierric Cistac, Tim Rault, Rémi Louf, Morgan Funtowicz, Joe
  Davison, Sam Shleifer, Patrick von Platen, Clara Ma, Yacine Jernite, Julien
  Plu, Canwen Xu, Teven~Le Scao, Sylvain Gugger, Mariama Drame, Quentin Lhoest,
  and Alexander~M. Rush,
\newblock ``Transformers: State-of-the-art natural language processing,''
\newblock in {\em Proc. \ EMNLP}, 2020, pp. 38--45.

\bibitem{godfrey1992switchboard}
John~J Godfrey, Edward~C. Holliman, and Jane McDaniel,
\newblock ``Switchboard: Telephone speech corpus for research and
  development,''
\newblock in {\em Proc.\ IEEE ICASSP}, 1992, vol.~1, pp. 517--520.

\bibitem{shriberg2001observations}
Elizabeth Shriberg, Andreas Stolcke, and Don Baron,
\newblock ``Observations on overlap: findings and implications for automatic
  processing of multi-party conversation,''
\newblock in {\em Proc.\ Interspeech}, 2001, pp. 1359--1362.

\bibitem{hu2021lora}
Edward~J Hu, Yelong Shen, Phillip Wallis, Zeyuan Allen-Zhu, Yuanzhi Li, Shean
  Wang, Lu~Wang, and Weizhu Chen,
\newblock ``{LoRA}: Low-rank adaptation of large language models,''
\newblock {\em arXiv preprint arXiv:2106.09685}, 2021.

\end{thebibliography}

\end{document}